\documentclass{article}

\usepackage{PRIMEarxiv}

\usepackage[utf8]{inputenc} 
\usepackage[T1]{fontenc}   
\usepackage{hyperref}       
\usepackage{url}            
\usepackage{booktabs}       
\usepackage{amsfonts}       
\usepackage{nicefrac}       
\usepackage{microtype}      
\usepackage{lipsum}
\usepackage{fancyhdr}       
\usepackage{graphicx}       
\graphicspath{{media/}}     
\usepackage[numbers,sort&compress]{natbib}
\usepackage{amsmath}

\title{A Dual-Use Framework for Clinical Gait Analysis: \\ Attention-Based Sensor Optimization and \\ Automated Dataset Auditing
}

\author{
  Hamidreza Sadeghsalehi \\
  Imperial College London \\
}

\begin{document}
\maketitle

\begin{abstract}
\textbf{Background:} Objective gait analysis using wearable sensors is critical for managing neurological and orthopedic conditions. However, optimizing sensor placement for specific clinical tasks and mitigating the impact of hidden dataset biases on artificial intelligence (AI) models remain significant challenges.
\textbf{Methods:} A multi-stream, attention-based deep learning model was applied to the Voisard et al. (2025) multi-cohort gait dataset, addressing four binary classification tasks: Parkinson's Disease (PD) screening, Osteoarthritis (OA) screening, post-stroke asymmetry detection (CVA), and a differential diagnosis (PD vs. CVA). We implemented strategies to mitigate severe class imbalance and evaluated the model using imbalance-robust metrics (e.g., Balanced Accuracy, MCC) and bootstrapped confidence intervals. The model's primary outputs were classification performance and a quantitative map of learned sensor importance weights for four body locations: Head (HE), Lower Back (LB), Left Foot (LF), and Right Foot (RF).
\textbf{Results:} The model's attention mechanism identified distinct, task-specific sensor synergies, such as a novel Head-Right-Foot (HE-RF) combination for PD screening and a Head-Left-Foot (HE-LF) combination for the complex PD-vs-CVA differential diagnosis. Critically, the model functioned as an automated data auditor by discovering a severe laterality confound in the dataset. For both OA and CVA screening—tasks where bilateral information is clinically essential—the model assigned over 70\% of its attention exclusively to the Right Foot sensor, with near-zero (e.g., <0.1\%) attention to the Left Foot. This counter-intuitive result, confirmed by statistically significant 95\% CIs, was a direct reflection of a severe laterality bias in the patient cohorts (15/0 right-sided OA; 47/2/0 right-dominant CVA) and not a valid clinical finding.
\textbf{Conclusion:} The attention-based framework is a powerful dual-use tool. Its primary methodological contribution is its function as an automated data auditor capable of discovering and quantifying hidden dataset biases, a critical step for responsible AI development. Concurrently, when applied to valid data subsets, it provides a data-driven method for identifying minimal, clinically relevant sensor sets (e.g., HE+RF for PD) to optimize future clinical monitoring protocols.
\end{abstract}


\section{Introduction}
Gait impairment is a cardinal symptom and a major contributor to disability across a wide spectrum of chronic health conditions, particularly in neurology and orthopedics \cite{mirelman2019gait,patterson2010evaluation,mills2013biomechanical}. In Parkinson's Disease (PD), characteristics such as reduced walking speed, shortened step length, and increased gait variability are not merely symptoms but are core to the disease's motor pathology and progression \cite{mirelman2019gait,del2016free}. Similarly, in conditions like osteoarthritis (OA) or following a cerebrovascular accident (CVA), alterations in gait symmetry and kinematics are primary indicators of disease severity and functional limitation \cite{mills2013biomechanical,patterson2010evaluation}. Traditional clinical assessment of gait, however, often relies on observational scoring and qualitative descriptions; while valuable, these methods can lack the sensitivity and objectivity required to track subtle changes over time, evaluate the efficacy of interventions, or develop precise diagnostic and prognostic models \cite{baker2007history,mancini2010relevance}. This gap has created a clear clinical imperative for quantitative, objective, and scalable methods for gait assessment.

The advent of wearable sensors, particularly inertial measurement units (IMUs), has catalyzed a paradigm shift in movement analysis \cite{tao2012gait,godfrey2008direct}. These low-cost, unobtrusive devices can capture high-fidelity kinematic data—including acceleration and angular velocity—in ambulatory, real-world settings, overcoming the spatial and temporal limitations of traditional laboratory-based, optical motion capture systems \cite{baker2007history}. The resulting high-dimensional time-series data streams offer a rich substrate for the development of digital biomarkers capable of objectively quantifying pathological gait patterns \cite{coravos2019developing}. Large-scale, clinically annotated datasets, such as the comprehensive collection provided by Voisard et al. (2025), are instrumental in bridging the gap between raw sensor data and translational clinical research, providing the necessary scale and diversity to develop robust analytical models \cite{voisard2025dataset}.

Despite these advances, a fundamental and often overlooked question persists: for a given clinical question, what is the minimal yet sufficient set of sensors required for an accurate assessment \cite{caramia2018imu,storm2016gait,van2019overview}? Current research and clinical protocols frequently employ either a large array of sensors, which can be burdensome for patients and computationally intensive, or rely on heuristic choices for sensor placement. This one-size-fits-all approach is suboptimal, as different pathologies manifest through distinct kinematic signatures across different body segments. For example, the systemic motor deficits of PD, which affect both axial trunk control and appendicular limb movement, are fundamentally different from the localized joint-level impairments of unilateral knee OA \cite{mirelman2019gait,mills2013biomechanical}. A logical and more efficient paradigm would therefore be task-dependent, tailoring the sensor configuration to the specific clinical question being asked.

Deep learning models, including Convolutional Neural Networks (CNNs) and Recurrent Neural Networks (RNNs), have emerged as state-of-the-art for analyzing complex time-series data from wearable sensors, demonstrating strong performance in tasks ranging from fall risk classification to gait recognition \cite{hammerla2016deep,ordonez2016deep,meyer2020wearables}. Within this domain, the attention mechanism, a technique originally developed for natural language processing, offers a powerful means of enhancing both model performance and interpretability \cite{bahdanau2014neural}. By enabling a model to dynamically assign importance weights to different segments of its input, attention can be adapted for multi-sensor fusion, learning the relative contribution of each sensor's data stream to a final prediction \cite{ma2019attnsense}. While the reliability of attention mechanisms for explaining individual predictions has been a subject of debate—as single-instance attention maps can be inconsistent across model initializations—their utility can be harnessed more robustly \cite{jain2019attention,serrano2019attention,wiegreffe2019attention,yadav2025attention}. By aggregating attention weights across an entire cohort, the mechanism can be transformed from a potentially fragile explanator into a stable and powerful tool for discovering aggregate, population-level insights about feature importance for a given task.

Parallel to the challenge of optimizing sensor configurations is the pervasive and critical problem of dataset bias in medical AI \cite{obermeyer2019dissecting,geirhos2020shortcut,zech2018variable,vayena2018machine,rajkomar2019machine,kelly2019key}. AI models are susceptible to learning and amplifying biases present in their training data, which can lead to poor generalization, reinforcement of health inequities, and clinically harmful outcomes \cite{obermeyer2019dissecting,zech2018variable}. Confounding variables—extraneous factors correlated with both input and outcome—can cause a model to learn spurious associations \cite{roberts2021common}. For instance, if a disease cohort is significantly older than a control cohort, a model may inadvertently become a highly accurate age detector rather than a true disease detector \cite{zech2018variable}. This underscores the urgent need for rigorous data auditing and governance as a foundational step in responsible AI development \cite{roberts2021common,kelly2019key,rajkomar2019machine,chakradeo2025navigating}. However, such auditing is typically a manual, resource-intensive process, ill-equipped to uncover subtle or unexpected confounders in large, complex datasets \cite{roberts2021common}. This contrasts with our work, which explores an emergent auditing capability embedded within the model itself.

This paper presents a dual contribution to the fields of digital health and medical AI. First, we introduce a novel, attention-based deep learning framework that provides a data-driven approach to identifying minimal, task-specific sensor configurations for clinical gait analysis. Second, and more significantly, we demonstrate that this same framework functions as a powerful tool for automated dataset auditing. We show how the model's learned attention weights can quantitatively expose a critical, hidden laterality confound within a major public dataset, transforming the model from a simple classifier into an engine for data quality assurance and scientific discovery.

\section{Methods}

\subsection{Study Dataset and Participants}
This study utilized the “Dataset of Clinical Gait Signals with Wearable Sensors,” a publicly available, multi-pathology, and clinically annotated dataset curated by Voisard et al. (2025) \cite{voisard2025dataset}. The dataset comprises 1356 gait trials from a total of 260 participants, organized into three primary groups: Healthy Subjects (HS), patients with neurological conditions (Neuro), and patients with orthopedic conditions (Ortho) \cite{voisard2025dataset}. For the specific tasks defined in this study, participants were selected from four cohorts: Healthy Subjects (HS, n=73), Parkinson's Disease (PD, n=143), Hip Osteoarthritis (HOA, n=44), and Cerebrovascular Accident (CVA, n=47) \cite{voisard2025dataset}. The original study was conducted in accordance with the Declaration of Helsinki, and all participants provided informed consent; full details regarding participant recruitment, clinical scoring, and ethical approvals are available in the original data descriptor \cite{voisard2025dataset}.

\subsection{Data Acquisition and Preprocessing}
Gait data were collected using four IMU sensors (XSens or Technoconcept) placed on the Head (HE), Lower Back at the L5 vertebra level (LB), Left Foot (LF), and Right Foot (RF) \cite{voisard2025dataset}. Each sensor provided a 9-channel time-series data stream, consisting of 3D accelerometer signals, 3D gyroscope signals, and 3D gravity-corrected free acceleration signals \cite{tao2012gait}. All data were sampled at a frequency of 100 Hz. The standardized data collection protocol involved participants performing a 10-meter walk, followed by a 180-degree U-turn and a 10-meter walk back to the starting point \cite{voisard2025dataset}. For this analysis, the preprocessed data files provided by Voisard et al. were used; these contain synchronized, multi-sensor time-series data for each trial, along with metadata including detected gait events such as heel-strikes and toe-offs \cite{voisard2025dataset,mariani2013quantitative}.

\subsection{Formulation of Clinical Classification Tasks}
To evaluate the model's ability to learn task-specific sensor configurations, four distinct binary classification tasks were formulated, each designed to simulate a relevant clinical question:
\begin{itemize}
  \item \textbf{PD Screening:} Differentiating between participants with Parkinson's Disease (PD) and Healthy Subjects (HS). This task simulates the use of gait analysis as a screening tool for a common neurodegenerative disorder \cite{caramia2018imu}.
  \item \textbf{OA Screening:} Differentiating between participants with Hip Osteoarthritis (HOA) and Healthy Subjects (HS). This task models the identification of gait patterns associated with a prevalent orthopedic condition \cite{mills2013biomechanical}.
  \item \textbf{Asymmetry Detection:} Differentiating between participants post-Cerebrovascular Accident (CVA) and Healthy Subjects (HS). Given the typically unilateral motor deficits following a stroke, this task was designed to test the model's ability to identify asymmetrical gait \cite{patterson2010evaluation}.
  \item \textbf{Differential Diagnosis:} Differentiating between participants with Parkinson's Disease (PD) and those with Cerebrovascular Accident (CVA). This represents a more complex clinical challenge, requiring the model to distinguish between two distinct pathologies that can both affect gait.
\end{itemize}

\subsection{Attention-Based Multi-Stream Neural Network Architecture}
A multi-stream neural network incorporating a sensor-level attention mechanism was designed to process the multi-modal sensor data \cite{bahdanau2014neural,ma2019attnsense}. The architecture consists of three main components:
\begin{itemize}
  \item \textbf{Modality-Specific Feature Extraction:} The model employs four parallel and independent branches, one for each sensor location (HE, LB, LF, RF). Each branch is a 1D Convolutional Neural Network (1D-CNN) that processes the 9-channel time-series input from its corresponding sensor. These CNNs are designed to learn high-level, abstract feature representations from the raw sensor signals. The output of each branch is a fixed-length feature vector, \(v_i \in \mathbb{R}^{128}\), where \(i \in \{\mathrm{HE}, \mathrm{LB}, \mathrm{LF}, \mathrm{RF}\}\).

  Each 1D-CNN branch consists of three sequential convolutional layers with filter counts of 32, 64, and 128, respectively. All layers use a \texttt{kernel\_size=15} and \texttt{padding=7}. Each convolutional layer is followed by a \texttt{ReLU} activation and a \texttt{MaxPool1d} layer with \texttt{kernel\_size=2}. A final \texttt{AdaptiveAvgPool1d(1)} layer collapses the time dimension, producing the 128-dimensional feature vector for each sensor.

  \item \textbf{Sensor-Level Attention Mechanism:} The four feature vectors \(\{v_{\mathrm{HE}}, v_{\mathrm{LB}}, v_{\mathrm{LF}}, v_{\mathrm{RF}}\}\) are fed into an attention module. This module consists of a single linear layer that learns to compute four unnormalized scalar importance scores, \(e_i\). These scores are then passed through a softmax function to generate the final attention weights, \(\alpha_i\), which represent the model's learned judgment of each sensor's relative importance for the given task. The weights are positive and sum to one.
  \[
    \alpha_i = \frac{\exp(e_i)}{\sum_{j \in \{\mathrm{HE}, \mathrm{LB}, \mathrm{LF}, \mathrm{RF}\}} \exp(e_j)}
  \]
  \item \textbf{Fusion and Classification:} The four feature vectors are fused into a single context vector via a weighted summation using their learned attention weights: \(c = \sum_i \alpha_i v_i\). This context vector, which represents a task-optimized summary of the information from all four sensors, is then passed through a final 2-layer fully-connected classifier (with \texttt{ReLU} activation and \texttt{Dropout(0.5)}) to produce a single logit for binary class prediction.
\end{itemize}

\subsection{Experimental Design and Statistical Analysis}
To ensure robust and generalizable evaluation, the dataset was split at the patient level into training (70\%), validation (15\%), and test (15\%) sets, a critical methodological step to prevent data leakage \cite{kaufman2012leakage,roberts2021common}. The exact class distributions (number of patients vs. controls) for each test set are reported in Table~\ref{tab:performance}.

For each of the four clinical tasks, a separate model was trained from a random initialization. Given the significant class imbalance observed in the data splits, models were trained using a weighted binary cross-entropy loss function (\texttt{nn.BCEWithLogitsLoss} with \texttt{pos\_weight}), where the positive class weight was set to $\frac{\text{negative\_samples}}{\text{positive\_samples}}$ to penalize misclassification of the minority class.

Training was performed using the Adam optimizer with a learning rate of \texttt{1e-4} for 50 epochs, a \texttt{batch\_size} of 32, and early stopping with a \texttt{patience} of 5 epochs based on validation loss.

Model performance was evaluated on the held-out test set using metrics robust to class imbalance, including: Area Under the Receiver Operating Characteristic Curve (ROC-AUC), Precision-Recall AUC (PR-AUC), Balanced Accuracy, Matthew's Correlation Coefficient (MCC), Sensitivity (Recall), and Specificity. Ninety-five percent confidence intervals for these metrics were calculated using 1000 bootstrap samples.

The primary explanatory output of the model, the ``Sensor Importance Map,'' was derived for each task by calculating the mean of the learned attention weights \((\alpha_{\mathrm{HE}}, \alpha_{\mathrm{LB}}, \alpha_{\mathrm{LF}}, \alpha_{\mathrm{RF}})\) across all samples in the test cohort. We calculated 95\% confidence intervals for these mean attention weights using 1000 bootstrap samples to provide a measure of statistical uncertainty in the model's learned strategy.

\section{Results}

\subsection{Performance on Imbalanced Clinical Tasks}
Our initial analysis revealed significant class imbalance across all tasks. Consequently, we adopted a weighted loss function and evaluated performance using imbalance-robust metrics, which are presented in Table~\ref{tab:performance}. The model demonstrated strong discriminative ability on the OA Screening (ROC-AUC 0.990) and Asymmetry Detection (ROC-AUC 0.950) tasks. Performance on the PD Screening (ROC-AUC 0.821) and the highly complex Differential Diagnosis (ROC-AUC 0.657) tasks was more modest, reflecting the clinical difficulty of these classifications.

The metrics highlight the challenge of class imbalance. For PD Screening, despite a high Specificity (0.945), the Sensitivity (0.333) and Balanced Accuracy (0.639) indicate the model still struggles to identify the minority PD class, even with re-weighting. Conversely, the model achieved perfect Sensitivity (1.000) for OA Screening, correctly identifying all positive-class patients in the test set. These results underscore the necessity of using metrics beyond simple accuracy to interpret model performance in clinical datasets.

\begin{table}[t]
  \centering
  \caption{Model classification performance and mean sensor attention weights on the test set. Performance is reported with metrics robust to class imbalance. Attention weights are shown as mean \% (95\% Confidence Interval).}
  \label{tab:performance}
  \resizebox{\linewidth}{!}{
  \begin{tabular}{l|c|ccccc|ccccc} 
    \toprule
    \textbf{Clinical Task} & \textbf{N (Pat./Ctrl.)} & \textbf{ROC-AUC} & \textbf{PR-AUC} & \textbf{Bal. Acc.} & \textbf{MCC} & \textbf{Sens. (Rec.)} & \textbf{Spec.} & \textbf{HE (\%)} & \textbf{LB (\%)} & \textbf{LF (\%)} & \textbf{RF (\%)} \\
    \midrule
    PD Screening & 15 (4 Pos/11 Neg) & 0.821 & 0.671 & 0.639 & 0.369 & 0.333 & 0.945 & 32.9 (28.3--37.4) & 3.4 (3.1--3.8) & 11.2 (10.3--12.4) & 52.5 (48.4--56.9) \\
    OA Screening & 14 (2 Pos/12 Neg) & 0.990 & 0.906 & 0.942 & 0.682 & 1.000 & 0.885 & 6.3 (4.9--7.7) & 22.7 (18.2--27.5) & 0.1 (0.0--0.1) & 71.0 (65.7--76.0) \\
    Asymmetry (CVA) & 19 (8 Pos/11 Neg) & 0.950 & 0.753 & 0.747 & 0.521 & 0.563 & 0.932 & 22.6 (17.1--28.1) & 0.0 (0.0--0.0) & 0.0 (0.0--0.0) & 77.4 (71.8--82.9) \\
    Differential (PD vs. CVA) & 11 (4 Pos/7 Neg) & 0.657 & 0.821 & 0.607 & 0.202 & 0.500 & 0.714 & 51.5 (41.8--60.6) & 0.2 (0.1--0.2) & 46.8 (37.9--56.4) & 1.5 (1.1--2.0) \\
    \bottomrule
  \end{tabular}
}
\end{table}

\subsection{Attention Mechanism Discovers Task-Specific Sensor Importance Maps}
The primary goal of the study was to determine if the model could learn to prioritize different sensors based on the clinical question. The aggregated attention weights, summarized in Table~\ref{tab:performance} and visualized in Figures~1--4, revealed distinct and complex sensor importance maps for each task.

\begin{figure}[t]
  \centering
  \includegraphics[width=0.7\textwidth]{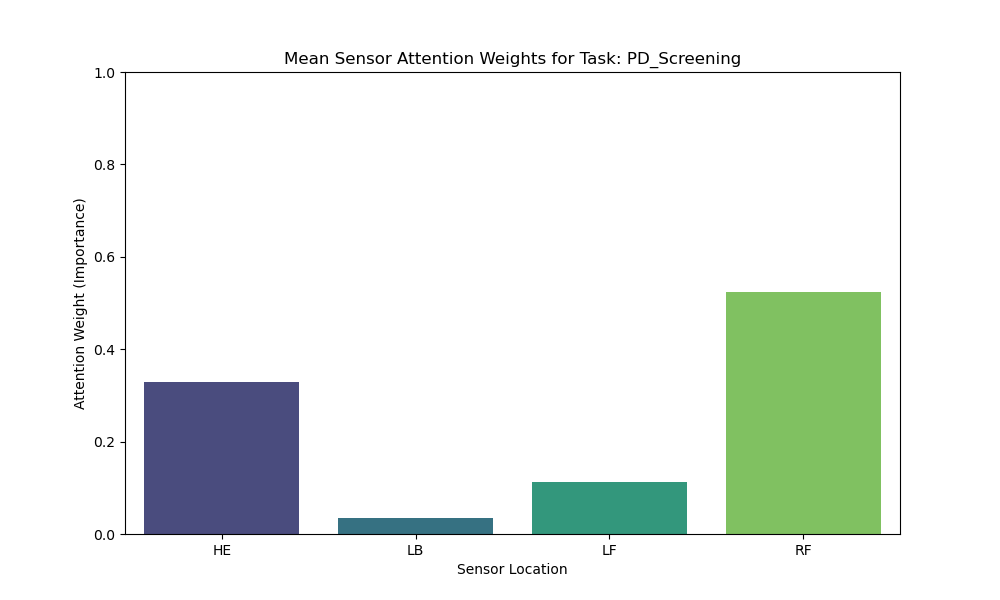}
  \caption{Mean Sensor Attention Weights for Parkinson's Disease (PD) Screening. The model learned to prioritize a novel synergy between the Right Foot (RF) (52.5\%) and Head (HE) (32.9\%) sensors. This data-driven hypothesis aligns with the clinical presentation of PD, which involves both appendicular (gait) and axial (postural) motor deficits. Error bars represent 95\% confidence intervals calculated from 1000 bootstrap samples, indicating the statistical uncertainty in the estimate of the mean attention.}
  \label{fig:pd_attention}
\end{figure}

\textbf{PD Screening (Figure~\ref{fig:pd_attention}):} For the PD Screening task, the model learned a novel synergy, allocating the most attention to the Right Foot (RF) sensor (52.5\%) and the Head (HE) sensor (32.9\%). This data-driven finding suggests a strategy of integrating information from both appendicular (foot) and axial (head) segments. This aligns well with the known pathophysiology of Parkinson's Disease, which is characterized by both appendicular motor deficits (e.g., shuffling, reduced step length) and axial symptoms (e.g., postural instability, head tremor) \cite{mirelman2019gait,del2016free}. This HE-RF combination is a plausible, data-driven hypothesis for an efficient PD screening protocol.

\begin{figure}[t]
  \centering
  \includegraphics[width=0.7\textwidth]{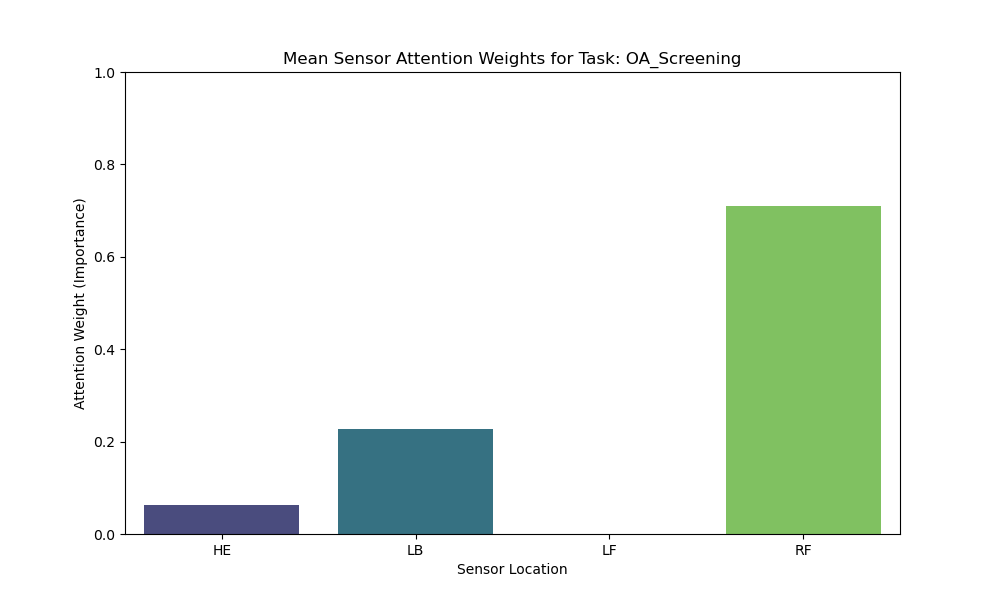}
  \caption{Mean Sensor Attention Weights for Osteoarthritis (OA) Screening. This result demonstrates the model's function as a data auditor. It assigned over 70\% of its attention to the Right Foot (RF) sensor while statistically ignoring the Left Foot (LF) (0.1\%). This counter-intuitive pattern reflects the model's discovery of a dataset confound (a 15/0 right-sided laterality bias in the HOA cohort), not a generalizable clinical signature of the disease. Error bars represent 95\% confidence intervals (1000 bootstrap samples).}
  \label{fig:oa_attention}
\end{figure}

\begin{figure}[t]
  \centering
  \includegraphics[width=0.7\textwidth]{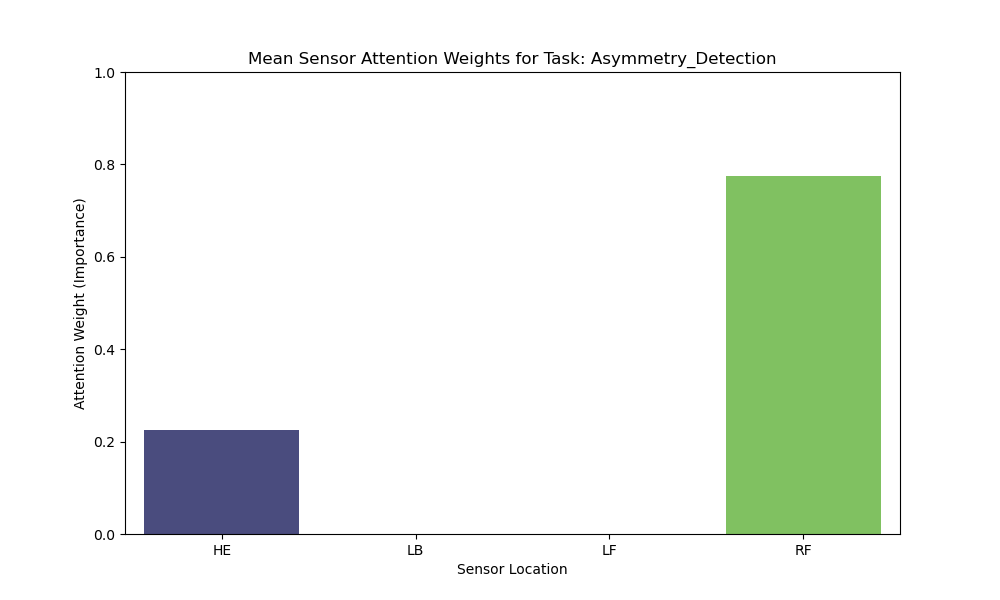}
  \caption{Mean Sensor Attention Weights for Asymmetry Detection (CVA). Similar to the OA task, the model relied almost exclusively on the Right Foot (RF) sensor (77.4\%). The lack of attention to the Left Foot (LF) (0.0\%) for a task predicated on asymmetry demonstrates the exploitation of a dataset confounder (a 47/2/0 right-dominant laterality bias in the CVA cohort), validating the attention mechanism's ability to flag hidden biases. Error bars represent 95\% confidence intervals (1000 bootstrap samples).}
  \label{fig:cva_attention}
\end{figure}

\textbf{OA and Asymmetry Screening (Figure~\ref{fig:oa_attention} \& \ref{fig:cva_attention}):} In stark contrast, the results for OA Screening and Asymmetry Detection (CVA) are not clinical findings, but rather a clear validation of our auditing methodology. For both tasks, the model developed an overwhelming and statistically significant reliance on a single sensor: the Right Foot (RF). In the OA task, the RF sensor received 71.0\% of the attention, and in the CVA task, it received 77.4\%.
Critically, the Left Foot (LF) sensor was actively and confidently ignored in both scenarios, with mean attention weights of 0.1\% and 0.0\%, respectively. The 95\% confidence intervals for the LF sensor attention (e.g., [0.0--0.1]) confirm that this is not an unstable result but a deliberate and stable strategy. This finding is highly counter-intuitive, as a clinical diagnosis of asymmetry or OA fundamentally relies on comparing bilateral limb information \cite{mills2013biomechanical,patterson2010evaluation}. The model's categorical rejection of the left foot strongly suggested it had discovered a dataset confound, which we confirm in Section 3.3.

\begin{figure}[t]
  \centering
  \includegraphics[width=0.7\textwidth]{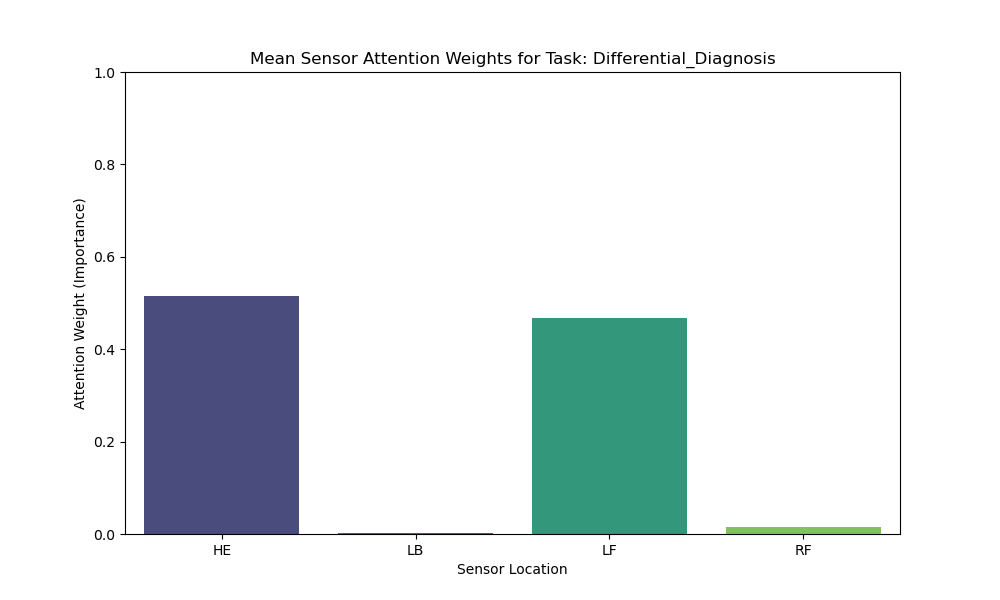}
  \caption{Mean Sensor Attention Weights for Differential Diagnosis (PD vs.\ CVA). To solve this complex task, the model learned a sophisticated synergy, allocating importance to the Head (HE) sensor (51.5\%) and the Left Foot (LF) sensor (46.8\%). It learned to ignore the confounded Right Foot sensor (1.5\%), suggesting a strategy of comparing central postural control (Head) with a non-confounded limb (Left Foot). Error bars represent 95\% confidence intervals (1000 bootstrap samples).}
  \label{fig:diff_attention}
\end{figure}

\textbf{Differential Diagnosis (Figure~\ref{fig:diff_attention}):} The most sophisticated sensor utilization strategy was observed in the Differential Diagnosis task. To distinguish between PD and CVA, the model learned a unique, non-obvious synergy, focusing almost equally on the Head (HE) sensor (51.5\%) and the Left Foot (LF) sensor (46.8\%). The Right Foot, which dominated the CVA screening task, was now almost completely ignored (1.5\%). This demonstrates a remarkable level of adaptive learning. The model appears to have learned that the RF sensor is a confounder for CVA and thus provides no utility in differentiating it from PD. Instead, it reasons that it must compare central postural control (Head, often affected in PD) with a non-confounded limb (Left Foot) to solve the task. The wide, overlapping confidence intervals for HE and LF weights also suggest a high-variance strategy, reflecting the nuanced clinical challenge of this task.

\subsection{Automated Discovery of a Critical Dataset Confounder}
The anomalous and counter-intuitive findings from the OA and CVA screening tasks prompted a deeper investigation into the dataset itself. The model's extreme and consistent preference for the right foot sensor was not a model failure but a successful and quantitative discovery of a significant sampling bias in the underlying Voisard et al. (2025) dataset.

A review of the cohort demographics detailed in the original data descriptor publication revealed a critical confound related to pathology laterality \citep{voisard2025dataset}:
\begin{itemize}
  \item \textbf{HOA Cohort:} The laterality of hip osteoarthritis was predominantly right-sided. The paper reports a right/left laterality of 15/0, meaning all patients with unilateral hip pathology had it on the right side.
  \item \textbf{CVA Cohort:} The cohort of stroke survivors was overwhelmingly composed of individuals with right-sided motor deficits. The reported laterality (right/left/ambidextrous) was 47/2/0, indicating that nearly all unilateral cases involved the right side of the body.
\end{itemize}
The model, as an optimization algorithm, learned the most parsimonious path to a correct classification. In this biased dataset, the most predictive feature for identifying a CVA or HOA patient was not a complex measure of ``asymmetry'' but simply the presence of an ``abnormality in the right foot's signal.'' The model correctly deduced that the Left Foot sensor provided redundant or non-informative data for these specific tasks and therefore assigned it a near-zero attention weight. This finding powerfully demonstrates the capacity of an interpretable model to function as an automated data auditor, flagging hidden confounders that could severely limit the generalizability of any model trained on such data.

\section{Discussion}
This study demonstrates the dual utility of an attention-based deep learning framework for wearable sensor-based gait analysis. The principal findings are twofold: first, the model successfully identified minimal, task-specific sensor configurations, providing data-driven hypotheses for optimizing clinical monitoring protocols. Second, and more significantly, the model's interpretability layer served as an effective, automated tool for discovering and quantifying a severe laterality confound in a major public dataset, highlighting a novel application for such architectures in data quality assurance and scientific discovery.

\subsection{Clinical Implications for Optimized Gait Analysis Protocols}
The sensor importance maps generated by our model offer actionable hypotheses for designing more efficient and less burdensome clinical gait analysis protocols.
For PD screening, the model's reliance on a Head-Right-Foot (HE-RF) synergy suggests that a minimal two-sensor setup, capturing both axial and appendicular features, could be a powerful data-driven hypothesis for future investigation \citep{caramia2018imu,del2016free}.
This has significant implications for remote patient monitoring and decentralized clinical trials, where patient adherence and ease of use are paramount \citep{coravos2019developing}.
For more complex tasks like differential diagnosis, the model's learned HE-LF synergy suggests that capturing information from anatomically distant body segments, reflecting both central and peripheral nervous system function, may be necessary.
These data-driven insights move beyond heuristic sensor placement, offering a principled approach to protocol design that is tailored to the specific clinical question at hand.

\subsection{Interpretable AI as an Engine for Automated Scientific Discovery and Data Auditing}
The most profound contribution of this work is methodological.
The discovery of the laterality confound was not a limitation of our study but its most powerful result.
It showcases a new paradigm for the use of interpretable AI: not merely as a tool for prediction, but as an active instrument for probing, validating, and understanding the datasets upon which our models are built.
While the broader literature on responsible AI rightly focuses on mitigating demographic and societal biases to ensure fairness and equity \citep{vayena2018machine,rajkomar2019machine,kelly2019key,obermeyer2019dissecting,geirhos2020shortcut,chakradeo2025navigating}, our work demonstrates that these same principles and tools can be extended to uncover clinical and sampling biases that are equally detrimental to a model's scientific validity and generalizability.
This automated approach represents a scalable and powerful alternative to the often manual and time-consuming processes of data auditing \citep{roberts2021common}.
An interpretable model can serve as a ``canary in the coal mine,'' quantitatively flagging potential confounders that might be missed by human reviewers examining summary statistics.
By training models on various sub-tasks within a dataset and analyzing their learned attention patterns, researchers can systematically screen for hidden biases, thereby strengthening the foundations of evidence-based digital medicine.

\subsection{Limitations and Considerations}
It is crucial to acknowledge the limitations of this study.
First, the specific sensor importance maps for the OA and CVA tasks are intrinsically linked to the Voisard et al. (2025) dataset and its identified confounder.
They should not be interpreted as generalizable gait signatures for these conditions but rather as signatures of the dataset's specific bias.
Second, the compelling sensor importance maps derived for PD Screening and Differential Diagnosis must be treated as data-driven hypotheses that require urgent validation on independent, multi-centric datasets. It is paramount that these validation datasets are themselves screened for similar biases (e.g., laterality) and other potential confounders not analyzed here, such as systematic differences in age, disease severity, or walking speed between cohorts, which could also be learned by the model.

Furthermore, our methodology must be contextualized within the active 'attention as explanation' debate \citep{jain2019attention,serrano2019attention,wiegreffe2019attention,yadav2025attention}. We acknowledge that instance-level attention weights can be unstable and may not reliably reflect a model's decision-making process for a single prediction. However, our approach deliberately avoids this contentious use case. Instead, we use cohort-level aggregation—averaging the attention weights across all samples in the test set—to identify stable, population-level importance maps. We contend this aggregation effectively averages out instance-level noise and instability, revealing a robust and meaningful signal about the model's learned strategy for the task as a whole. This aggregated map, supported by bootstrapped confidence intervals, reflects the underlying data distribution and feature importance, transforming attention from a fragile 'explanator' into a reliable tool for dataset auditing.

Finally, our choice of a CNN-based architecture is one of many valid approaches for time-series analysis; other architectures, such as Transformers, might yield different performance or insights \citep{zerveas2021transformer,xu2023transehr}.

\subsection{Future Directions}
This work opens several avenues for future research. The immediate next step is the prospective validation of the hypothesized minimal sensor sets (e.g., the HE-RF configuration for PD monitoring) in a real-world clinical study to confirm their efficacy and clinical utility. Beyond this, the methodology can be expanded into a formal, generalized framework for AI-driven dataset auditing. Such a framework could be deployed as a standard pre-analysis step in digital health research to automatically screen for potential confounders.

Once a confounder is identified, future work should focus on its mitigation. This could involve integrating advanced techniques designed to train confounder-free models, such as the adversarial de-confounding methods cited in \cite{zhao2020training}, to explicitly remove the influence of the discovered laterality bias from the learned feature representations.
The long-term vision is the integration of these validated, optimized, and bias-aware models into clinical decision support systems and remote patient monitoring platforms, enabling more precise, efficient, and equitable care.

\section{Conclusion}
The attention-based deep learning framework presented in this study provides a robust, data-driven method for generating hypotheses about sensor optimization in clinical gait analysis. However, its true novelty and most significant contribution lie in its emergent capacity as an automated data auditor. By revealing hidden scientific insights and critical dataset flaws through its interpretable architecture, this methodology demonstrates a powerful dual-use capability. It represents a significant step forward in our ability to build more efficient, reliable, and responsible AI systems for the future of digital health.

\section{Data Availability}
The ``Dataset of Clinical Gait Signals with Wearable Sensors'' is publicly available at the Scientific Data page: \url{https://www.nature.com/articles/s41597-025-05959-w}. For citation and resolution, use the DOI link: \url{https://doi.org/10.1038/s41597-025-05959-w}.

\section{Code Availability}
The complete code base used for model training, evaluation, and generation of all results presented in this paper is available at: \url{https://github.com/hamidreza-s-salehi/GaitSensorAttention}.


\bibliographystyle{unsrt}
\bibliography{references}

\end{document}